\documentclass[manuscript]{acmart} 

\AtBeginDocument{%
  \providecommand\BibTeX{{%
    \normalfont B\kern-0.5em{\scshape i\kern-0.25em b}\kern-0.8em\TeX}}}

\copyrightyear{2023}
\acmYear{2023}
\setcopyright{acmlicensed}
\acmConference[FAccT '23]{2023 ACM Conference on Fairness, Accountability, and Transparency}{June 12--15, 2023}{Chicago, IL, USA}
\acmBooktitle{2023 ACM Conference on Fairness, Accountability, and Transparency (FAccT '23), June 12--15, 2023, Chicago, IL, USA}
\acmPrice{15.00}
\acmDOI{10.1145/3593013.3594068}
\acmISBN{979-8-4007-0192-4/23/06}

\usepackage{framed}
\usepackage{amsthm}                 
\usepackage{amsfonts}               
\usepackage{mathtools}              
\usepackage[shortlabels]{enumitem}	

\newcommand{\ECE}{\mathsf{ECE}}
\newcommand{\ACE}{\mathsf{ACE}}
\newcommand{\MCE}{\mathsf{MCE}}

\newcommand{\Cvara}{\mathsf{CVaR}_\alpha}

\newcommand{\MLCE}{\mathsf{MLCE}}
\newcommand{\Db}{\mathsf{D}}
\newcommand{\Rb}{\mathsf{R}}

\newcommand{\XX}{\mathcal{X}}
\newcommand{\ZZ}{\mathcal{Z}}
\newcommand{\YY}{\mathcal{Y}}
\newcommand{\PP}{\mathcal{P}}
\newcommand{\LL}{\mathcal{L}}
\newcommand{\DD}{\mathcal{D}}
\newcommand{\II}{\mathcal{I}}

\newcommand{\Tcal}{\mathcal{T}}
\newcommand{\Hcal}{\mathcal{H}}
\newcommand{\Gcal}{\mathcal{G}}
\newcommand{\qC}{\mathcal{Q}_C}

\newcommand{\RR}{\mathbb{R}}
\newcommand{\NN}{\mathbb{N}}
\newcommand{\EE}{\mathbb{E}}
\newcommand{\var}{\mathbb{V}}

\newcommand{\ph}{\hat{p}}
\newcommand{\Ph}{\hat{P}}
\newcommand{\ptil}{\Tilde{p}}
\newcommand{\ptr}{p}

\newtheorem{proposition}{Proposition}
\newtheorem{definition}[proposition]{Definition}
\newtheorem{example}[proposition]{Example}

\newcommand{\highlight}[1]{\vspace{5pt}
 \par\noindent
 \colorbox{gray!10}{%
 \parbox{\linewidth-8pt}{
 \vspace{-4pt}
 #1
 \vspace{-4pt}}
 }
}

\renewcommand{\highlight}[1]{
 #1
}

\begin{document}

\title{On the Richness of Calibration}

\author{Benedikt Höltgen}
\email{benedikt.hoeltgen@uni-tuebingen.de}
\orcid{0009-0008-0882-9821}
\affiliation{%
  \institution{University of Tübingen}
  \streetaddress{Maria-von-Linden Strasse 6}
  \city{Tübingen}
  \country{Germany}
  \postcode{72076}
}

\author{Robert C. Williamson}
\email{bob.williamson@uni-tuebingen.de}
\orcid{0000-0002-8862-1412}
\affiliation{%
  \institution{University of Tübingen and Tübingen AI Center}
  \streetaddress{Maria-von-Linden Strasse 6}
  \city{Tübingen}
  \country{Germany}
  \postcode{72076}
}

\renewcommand{\shortauthors}{Höltgen and Williamson}

\begin{abstract}
Probabilistic predictions can be evaluated through comparisons with observed label frequencies, that is, through the lens of calibration. Recent scholarship on algorithmic fairness has started to look at a growing variety of calibration-based objectives under the name of multi-calibration but has still remained fairly restricted. In this paper, we explore and analyse forms of evaluation through calibration by making explicit the choices involved in designing calibration scores. We organise these into three grouping choices and a choice concerning the agglomeration of group errors. This provides a framework for comparing previously proposed calibration scores and helps to formulate novel ones with desirable mathematical properties. In particular, we explore the possibility of grouping datapoints based on their input features rather than on predictions and formally demonstrate advantages of such approaches. We also characterise the space of suitable agglomeration functions for group errors, generalising previously proposed calibration scores. Complementary to such population-level scores, we explore calibration scores at the individual level and analyse their relationship to choices of grouping. We draw on these insights to introduce and axiomatise fairness deviation measures for population-level scores. We demonstrate that with appropriate choices of grouping, these novel global fairness scores can provide notions of (sub-)group or individual fairness.
\end{abstract}




\maketitle

\section{Introduction}


Enabled by the now widespread availability of data and compute, many data-driven tasks are increasingly aided by machine learning systems.
Examples of this include the assessment of individual risk, for example for credit default.
This comes with its own risks such as structurally biased data collection \citep{Oneil2017}, reinforcement of power structures \citep{Kasy2021} or algorithmic monoculture \citep{Oneil2017, Kleinberg2021}.
It does, however, have the advantage that algorithms can be thoroughly assessed, also with regard to notions of fairness.
As \citet{Vredenburgh2022} argues, fairness is a standard of justice (among others) that can be defined as the requirement of 'respecting people’s claims in proportion to the strength of those claims'.
In this work, we concentrate on fair probabilistic predictions \textit{in light of given data} -- in particular, on notions of fairness based on calibration.
This is somewhat restrictive but -- contrary to more 'substantive' notions of fairness \citep{Green2022} -- something formal approaches are well suited for;
while there is a place for formal work on fairness, it cannot stand alone \citep{Selbst2019}.


The term 'calibration' can refer to either the agreement of predictions with the frequency of observed labels or the process of aligning them; we are concerned with and use it in the former sense.
\citet{Vredenburgh2022} notes that '[m]any computer scientists, social scientists, and philosophers favour calibration as the best metric for fairness'.
To sensibly compare predictions with label frequencies, it is necessary to consider datapoints in groups.
The most common choice for this are bins of the prediction space $[0,1]$, with some variations especially in recent years -- sometimes in the context of fairness and then often under the name of multi-calibration \citep{Hebert-Johnson2018}.
A step in the direction of local notions of calibration has been made in \citep{Luo2022} and \citep{Cabitza2022}, each proposing a global score with components that can be localised in a learned representation space and in prediction space, respectively.
An ethical analysis of calibration-based notions of fairness requires an understanding of the choices involved in arriving at such notions.
However, these choices have hardly been explicitly discussed or analysed in previous literature.
In this work, we present a conceptual framework from the ground up with the goal of generalising previous notions of \hbox{(multi-)}calibration and making relationships between them more explicit.
Contrary to the understanding of calibration on subgroups as a \textit{refinement} of group calibration, we demonstrate that calibration originates at a localised level and is then extended to groups.


We elaborate on the emergence of groups as a fundamental aspect of calibration in practice and on the historical development of its now prevalent understanding in Section~\ref{s:local_cal}.
A major goal of this work is then to point to the vast space of possible alternatives and analyse the choices involved.
We demonstrate advantages and disadvantages of such alternatives for assessing prediction quality and fairness.
Where possible, we capture salient aspects of these trade-offs mathematically.
We show that calibration allows for both global and local evaluation, that is, both at the population level and localised to some point in input space.
The former involves the additional choice of an agglomeration function over group calibration errors.
The choices involved in global calibration scores as well as the structure of the present paper are captured in the following diagram:

\[
  \underbrace{\text{principle\quad size\quad structure}}_\text{\normalsize grouping}\quad \text{agglomeration}
\]
Most fundamental is the choice of a general \textit{principle} according to which datapoints are grouped together.
This can be arbitrary but will usually be some notion of vicinity in input space or prediction space (Section~\ref{ss:gr_vicinity}).
Independent of the grouping principle, there is a trade-off concerning the \textit{size} of the group between robustness and informativeness (Section~\ref{ss:gr_size}).
The third choice involved in the selection of a grouping concerns its general \textit{structure}.
Here, the options include bins of fixed size but also overlapping groups of nearest neighbours or kernels (Section~\ref{ss:gr_structure}).
The remaining choice involved in a global score is the \textit{agglomeration} function applied to the group errors.
We show that the commonly used maximum and average of absolute group errors are the two extreme cases of a broader class of functions suitable for evaluating prediction quality (Section~\ref{s:scores}).
To introduce population-level fairness scores based on calibration, we then axiomatise a matching class of deviation measures (Section~\ref{ss:fair_agglomeration}).
Our analysis of the richness of possible choices culminates in a demonstration that calibration errors on suitable groups, agglomerated by a deviation measure, can provide scores that capture notions of either individual or (sub-)group fairness (Section~\ref{ss:fair_notions}).
%
%
%
We summarise our contributions as follows:
\begin{itemize}
    \item We show that calibration can be construed more broadly than often assumed, which allows novel approaches to evaluate probabilistic predictions. 
    \item We organise the space of possible calibration scores through a small number of largely independent choices, pointing out considerations and trade-offs relevant to these choices.
    We capture salient aspects of these trade-offs mathematically.
    \item We propose the (to our knowledge) first provably consistent estimators of individual calibration, based on kernels and $k$-nearest neighbours.
    \item We characterise the space of functions suitable for agglomerating calibration errors into global scores of prediction quality, capturing existing notions of calibration.
    \item We propose and axiomatise agglomeration functions for global calibration-based fairness scores and show that through suitable choices of grouping, such scores can provide notions of (sub-)group fairness as well as notions of individual fairness.
\end{itemize}


\section{Calibration From the Ground Up}
\label{s:local_cal}

A common definition of (perfect) calibration is that for all $v \in [0,1]$, among the datapoints which receive the prediction $v$, the (expected or observed) frequency of the label '1' is equal to $v$.
In this section, we motivate our perspective on calibration as a notion that actually originates at the individual level and requires choices of grouping in practice.
We also provide a tentative historical explanation for why the understanding of calibration is usually more narrow.
Throughout the paper, we investigate the problem of predicting binary labels $\{0,1\}$ from inputs in a space $\XX$.


\subsection{From individuals to groups}

Given access to a dataset $\{(x_i, y_i)\}_{i \in \NN}$ with infinite datapoints for each input $x \in \XX$,\footnote{Within this particular paragraph, and only here, we assume $\XX$ to be countable for ease of presentation.} a probabilistic predictor $\ph: \XX \to [0,1]$ of binary labels would be 'correct' if it predicts the true probability everywhere, as reflected in the limiting frequency of the label '1'.\footnote{Note that in general, such a limit need not exist. While the law of large numbers is often taken to show that such limits exist and do reflect the true probability with probability one, this depends on the particular measure imposed on the infinite-dimensional space of possible sequences \citep[p. 8]{Schnorr2007}. We generally do not assume the existence of a true probability distribution in this paper, except where explicitly stated.}
Thus, a probabilistic predictor $\ph$ should satisfy
\begin{equation}\label{eq:infin_freq}
    \forall x \in \XX:\quad \ph(x) = \lim_{n \to \infty} \frac{1}{\lvert V_x^n\rvert} \sum_{i \in V_x^n} y_i,
\end{equation}
where
\begin{equation}
    V_x^n:= \{i \leq n: x_i=x\}.
\end{equation}
In practice, of course, we do not have access to infinite data.

From now on, we assume a finite dataset $\DD := \{(x_i, y_i)\}_{i \in \II}$ with $\II := \{1,...,N\}$ and $\XX_\DD := \{x_i\}_{i \in \II} \subset \XX$.
If we had lots of data for every input, we could still impose a criterion of perfect point-wise, 'individual' calibration. 
Now based on limited data, this would take the similar form
\begin{equation}\label{eq:fin_freq}
    \forall x \in \XX_\DD:\quad \ph(x) = \frac{1}{\lvert V_x\rvert} \sum_{i \in V_x} y_i,
\end{equation}
where
\begin{equation}\label{eq:fin_freq_V}
    V_x:= \{i \in \II: x_i=x\}.
\end{equation}
The availability of such abundant data would arguably render the deployment of machine learning models superfluous and suggest that the representation of individuals in the data is too coarse to treat them adequately.

In interesting settings, we have at most a few datapoints per input, and usually no data at all for some inputs.
This is, of course, why we need machine learning algorithms with constrained hypothesis classes and inductive biases in the first place.
To compare label frequencies with predictions 'at $x$', we then need to somehow extend the considered groups from those consisting only of identical inputs $V_x$ to more general groups of datapoints $I_x \subset \II$.\footnote{To make the most of the available information and avoid dependence on further sub-selection, it is desirable to always take \textit{all} datapoints that instantiate a given input, i.e. $I_x = \bigcup_{z \in U_x} V_{z}$ for some suitable subset of the input space of $U_x \subset \XX$.\label{fn:all_datapoints}}
This then leads to the more general criterion of calibration
\begin{equation} \label{eq:fin_calib}
    \forall x \in \XX_\DD:\quad \frac{1}{\lvert I_x\rvert} \sum_{i \in I_x} \ph(x_i) = \frac{1}{\lvert I_x\rvert} \sum_{i \in I_x} y_i.
\end{equation}
Note that, in contrast to (\ref{eq:infin_freq}) and (\ref{eq:fin_freq}), this does not suggest an 'optimal' value of $\ph(x)$.
Instead, it provides a quality criterion that can be localised to a particular point in input space. As Dawid notes, '[t]he calibration criterion has some similarity with the frequency definition of probability, but does not require a background of repeated trials under constant conditions' \citep[p. 606]{Dawid1982}.
Based on this, we can define the (signed, i.e. positive or negative) \textit{calibration error} of a group $I \subset \II$ to be
\begin{equation} \label{eq:cal_error}
    c(I) := \frac{1}{\lvert I \rvert} \sum_{i \in I} (\ph(x_i) - y_i).
\end{equation}


\subsection{Conventional understanding of calibration}

A lot of work on calibration can be seen as defining the $I_x$ via pre-defined bins $B_1, ..., B_K$ forming a partition of $[0,1]$ \citep{Zadrozny2001}.
The most common example of this is the Expected Calibration Error (ECE; \cite{Naeini2015}) defined as
\begin{equation}\label{eq:ece}
    \ECE(\ph) = \sum_{k=1}^K \frac{|B_k|}{N} \left\lvert \sum_{i: \ph(x_i) \in B_k} \ph(x_i) - y_i \right\rvert.
\end{equation}
While often assumed to be \textit{the} way to think about calibration, such prediction-based grouping is only one very specific choice, and not always suitable.
It does have the advantage that it allows for quickly assessing general under- or overconfidence of predictors, a question that has proven to be particularly relevant for modern neural networks \citep{Guo2017}.
For a formulation of the ECE (and other calibration scores) more in the style of (\ref{eq:fin_calib}), see Appendix~\ref{app:established_scores}.

However, there also seem to be contingent historical reasons for the prevalence of prediction-based grouping for calibration.
The evaluation of probabilistic predictors was primarily researched in the meteorological community and published in meteorology journals until well into the 70s, the Brier score \citep{Brier1950} being perhaps the first notable work.
Here, predictions were made by human forecasters -- potentially with the help of automated procedures \citep{Sanders1963} -- and their evaluations only had access to predictions and observed outcomes, not to the inputs.\footnote{Interestingly, it seems that these forecasters used a strategy of '\textit{sorting} all instances into an ordered set of categories of likelihood of occurrence, and \textit{labelling} each category with a specific likelihood, or probability, of occurrence' \citep[p. 200, original emphasis]{Sanders1963}. This fits nicely with then assessing the prediction quality through calibration based on groups of similar \textit{prediction}.}
Furthermore, these forecasters only assigned probabilities on a finite scale, such as multiples of 10\%, which did not require to impose bins and, thus, made these groups seem even more 'natural'.
Calibration in the sense of (\ref{eq:infin_freq}) - (\ref{eq:fin_calib}) was often called 'reliability' or 'validity', while 'calibration' was used in relation to over-and under-confidence \citep{Sanders1963}, closer to the original meaning of the word (regarding the grounding of scales or instruments).

When statisticians subsequently picked up the topic (and seem to have renamed it 'calibration'), they continued to use weather forecasting as the primary example and modelled the setup as a sequence of forecasts and outcomes.
They also extended the notion of calibration to basically the general setting of (\ref{eq:fin_calib}), speaking of subsequences rather than groups: \citet{Dawid1982} admits roughly any subsequence that does not take into account the ground-truth labels.
In more recent work, \citet{Dawid2017} suggests that also the information on which predictions are based can be taken into account for '$\Hcal$-based' grouping, but the discussion remains on an abstract level.

Roughly concurrently, work on 'multi-calibration' has suggested explicitly taking into account input information for selecting groups:
The original paper on multi-calibration \citep{Hebert-Johnson2018} proposed to look at all computable groups -- a proposal that has also been made in the sequence setting by \citep{Dawid1985}.
But even approaches based on similarity of learned representations either \textit{prescribe} that the selected groups have equal predictions \citep{Burhanpurkar2021} or apply additional binning in prediction space \citep{Luo2022}.\footnote{A very recent exception is \citep{Kelly2022}, suggesting to only use ECE-style binning on a single input feature like age.}
Optimisation targets that omit this additional binning are sometimes referred to as 'multi-accuracy' \citep{Hebert-Johnson2018, Kim2019} but we are not aware of a thorough discussion of this choice.
We find this name confusing since a predictor could be very wrong on every datapoint but still 'multi-accurate' -- this seems to not be in line with established notions of accuracy.\footnote{Consider a setting where in each group, there are equally many $1$s as $0$s among the labels and the predictor always predicts $\ph(x_i) = 1-y_i$ or always predicts $0.5$. In both cases, the average label per group, 0.5, coincides with the average prediction in every group.}
By taking into account the available structure in the input space, different notions of vicinity can be used to construct groups $I_x$ 'around $x$'.
Calibration errors of such groups allow to recover a localised sense of calibration -- we return to this in sections \ref{ss:gr_structure} and \ref{ss:fair_notions}.
In the following section, we discuss considerations concerning the choice of grouping. 

%
%

\FrameSep4pt
\begin{framed}
\noindent{\textbf{Intuition (Section~\ref{s:local_cal}):}
    To assess a predictor $\ph$ at a point $x$, we could compare it to the average label at $x$ if we had an infinite (\ref{eq:infin_freq}) or a large (\ref{eq:fin_freq}) number of datapoints for $x$.
    Since we usually do not have many datapoints for a given $x$, we can instead take a group of datapoints $I_x$ including it and compare their \textit{average} prediction to their ratio of positive labels (\ref{eq:fin_calib}).
    For partly historical reasons, $I_x$ has almost exclusively been taken to be the group of datapoints that is similar to $x$ in their \textit{predictions}.}
\end{framed}


\section{Grouping Choices}
\label{s:grouping}

Rather than a requirement, assessing calibration on prediction-based bins is only one particular choice.
This naturally leads to the question of what other options there are to group datapoints.
In the following, we constrain ourselves to groupings based on vicinity in the input- or prediction space, that is, based on notions of cohesion and similarity.
One alternative, suggested in \citep{Burhanpurkar2021}, would be to consider all efficiently computable groups.
Vicinity-based groups already give a wide space to explore and make groups and their errors particularly interpretable, which is helpful both for improving predictors and for questions of fairness.
Note that one may also pursue different approaches simultaneously to gain more insights, given that there is no 'correct' way of grouping.
In the words of Nelson Goodman, 'similarity is relative and variable, as undependable as indispensable' \citep[p. 20]{Goodman1972}.

To better understand a large space of possible choices, it is often helpful to consider inherent trade-offs.
For the case of grouping datapoints, these include vicinity in the input space versus in prediction space (Section~\ref{ss:gr_vicinity}) as well as the size of groups (Section~\ref{ss:gr_size}).
The third choice concerns the grouping structure, that is, how a notion of vicinity is applied to form groups (Section~\ref{ss:gr_structure}).


\subsection{Vicinity-based grouping principles}
\label{ss:gr_vicinity}

Notions of vicinity particularly relevant to grouping datapoints are those of group cohesion and individual similarity.
We formalise cohesion as a property of topological spaces, namely connectedness.
Note that we review basic notions of general topology in Appendix~\ref{app:topology}.
Groupings based on similarity make use of distances between points, e.g. through a metric.
In the sense that every metric space induces a topological space and every ball is a connected set in the induced topology but not vice versa, cohesion is weaker than similarity.
Cohesion is usually not sufficient to define groups, which means that vicinity-based groups generally rely on some notion of similarity.

A choice inherent in building groups based on vicinity is whether to focus on predictions $\ph(x_i)$ (as in the ECE) or input features $x_i$.\footnote{Interestingly, similarity of input and of prediction can, respectively, be approximately mapped to what \citet{Campbell1958} identified as the two conditions for 'entitativity' of groups: similarity and common fate.}
As described in the previous section, the popularity of prediction-based grouping may be partly due to the roots of calibration in meteorology where no useful structure in the input space was available for evaluating predictors.
Its main advantage is that it facilitates the assessment of over- and under-confidence (in terms of distance from $0.5$), by separating datapoints with high predictions from those with low predictions.
On the other hand, grouping datapoints based on vicinity in $\XX$ allows evaluating $\ph$ in specific areas of the input space. 
This can be desirable, for example, in the context of fairness where we are interested in the performance of $\ph$ on groups that are often in some way local or similar in the input space.
Some recent approaches propose to group points based on similarity of representations learned by some feature extractor \citep{Burhanpurkar2021, Luo2022}
An advantage of such approaches is that they implicitly make use of both inputs and predictions while a disadvantage is that they do so in an opaque way.
Especially for fairness applications, this might be difficult to justify since the ethical significance of such groups is less clear compared to groups defined explicitly in input or prediction space.

One important aspect of choosing a notion of vicinity is the mathematical structure that needs to be assumed.
Approaches to group fairness and multi-calibration mostly assume some set structure, often a partition given by individual categorical features (e.g. race intersecting with gender).
Relying on $\XX$-similarity requires that a sensible notion of distance such as a metric is available in the input space.\footnote{Strictly speaking, $k$-nearest neighbour approaches only require a complete and transitive ordering for every datapoint.}
However, we are not usually given a metric that captures similarity in a principled way. 
One option would then be to start with a standard metric (e.g. based on $\| \cdot \|_p$) and weight each feature dimension based on its observed variance or range, in an attempt to give each dimension equal relevance for the assessment of similarity.
Note that the exact form of the metric is not as central here as for Lipschitz-based notions of individual fairness; however, although it here only indirectly influences calibration scores via the grouping, it still remains a concern.
In this sense, $\XX$-similarity requires more additional structure than prediction similarity since the interval $[0,1]$ comes with much less controversy about the choice of metric:
All $\| \cdot \|_p$-based metrics induce the same intervals in $[0,1]$ and there is no issue of balancing multiple dimensions.
While there might be contexts where e.g. distances between points close to $0$ or $1$ should be elevated as compared to points close to $0.5$, the metric $|v - v'|$ seems generally well-suited to capture prediction similarity.
Also note that groupings based on learned representation implicitly use structure in both $\XX$ and $[0,1]$, through inductive biases of the extractor.

A trade-off between notions of vicinity arises due to the fact that disconnected areas in the input space may be mapped to the same predictions (or, analogously, the same representations), leading to different notions coming apart.
We characterise this tension in the language of topology, as motivated in Appendix~\ref{app:topology}. 
The proofs for all propositions can be found in Appendix~\ref{app:proofs}.
\highlight{
\begin{proposition}[Cohesion in domain and  codomain] \label{prop:cohesion} \ 
\begin{enumerate}[a)]
    \item[a)] For any continuous function $f: \XX \to \ZZ$, connectedness in $\XX$ implies connectedness in $\ZZ$ but not vice versa.
    \item[b)] For a non-constant predictor $\ph : \XX \to [0,1]$ with the standard topology on $[0,1]$, there is a topology on $\XX$ s.t. $\ph$ is not continuous (and thus connectedness in $\XX$ does not imply connectedness in $[0,1]$).
    \item[c)] If $\ph: \XX \to [0,1]$ has more than a single local minimum or more than a single local maximum, there exist intervals in $[0,1]$ whose preimages in $\XX$ are not connected (in any Hausdorff topology on $\XX$).
\end{enumerate}
\end{proposition}
}
%

Note that the placeholders $f$ and $\ZZ$ in a) can be seen as $\ph$ and $[0,1]$, respectively, but also as a feature extractor and its output space.
Groups of interest for fairness considerations usually depend on input features rather than predictions, such as groupings along lines of age, gender, race, and/or socio-economic background.
Proposition~\ref{prop:cohesion} shows that unless $\ph$ is very restricted, prediction-based grouping can lead to common assessment (in terms of a single calibration error) of areas in $\XX$ that are not connected.
Part c) in particular shows that only very simple predictors, essentially monotonous ones, avoid that some its approximate level sets (preimages of intervals in $[0,1]$) are disconnected.
A potentially problematic case in the context of fairness is when a small minority group (assumed to be similar to each other in $\XX$) have predictions similar to another, larger group such that the predictor's behaviour on the small group is washed out in the error of the combined group.
Another problem, not directly related to topology and relevant beyond the fairness context, is that of 'stereotyping', i.e. when predictions are similar in two (potentially adjacent) areas of $\XX$ in such a way that $\ph$ over-predicts in one but under-predicts in the other: this may suggest perfect calibration in the combined group as its average prediction may still coincide with the average label (cf. Proposition~\ref{prop:to_size_a}).
$\XX$-based grouping can detect such stereotyping while prediction-based grouping cannot.
Such problems are the reason why scholars have proposed to test calibration in multiple subgroups -- however, this still allows for such effects within these subgroups.



\subsection{Group size}
\label{ss:gr_size}

The trade-off between vicinity in input space and in prediction space cannot be resolved by arbitrarily intersecting $\XX$-similar with $\ph$-similar groups.
This is one implication of the more obvious trade-off concerning the size of the groups:
Larger groups allow more stable estimates while errors of smaller groups contain more information about their members.\footnote{\citet{Dawid2017} mentions a related but slightly different trade-off between robustness and incisiveness w.r.t. the amount of information taken into account for groups on which to calibrate forecasts.}
We formalise and quantify this in the following.
\highlight{
\begin{proposition}[Advantage of smaller groups: Resolution] \label{prop:to_size_a} \ \\
Take two nonempty, disjoint sets of indices $I_1, I_2 \subset \II = \{1,...,N\}$ and recall the definition of the calibration error in (\ref{eq:cal_error}).
Then 
\begin{enumerate}[a)]
    \item[a)] Assume that $\forall (i,j) \in I_1 \times I_2 : x_i \neq x_j$ and that not all labels $y_i, i \in I_1 \cup I_2$ are equal.
    Then a predictor $\ph: \XX \to [0,1]$ can be perfectly calibrated on $I_1 \cup I_2$ but neither on $I_1$ nor on $I_2$.
    \item[b)] For any predictor $\ph: \XX \to [0,1]$, $|c(I_1)| \leq \alpha$ and $|c(I_2)| \leq \beta$ entails $|c(I_1 \cup I_2)| \leq \frac{\alpha \cdot |I_1| + \beta \cdot |I_2|}{|I_1| + |I_2|}$.
    For $\alpha = \beta$, this means $|c(I_1 \cup I_2)| \leq \alpha$.
\end{enumerate}
\end{proposition}
}
%
%
%

\highlight{
\begin{proposition}[Advantage of larger groups: Variance] \label{prop:to_size_var} \ \\
Let $(X_1, Y_1), ..., (X_K, Y_K)$ be pairs of i.i.d. random variables with values in $\XX \times \{0,1\}$ and joint distribution $P$. 
Denote the Bayes-optimal predictor by $p(x) := \EE_P[Y | X=x]$ and take some predictor $\ph: \XX \to [0,1]$. 
Then
\begin{minipage}{\linewidth}
\vspace{4pt}
\begin{equation}
    \var_P \left[\frac{1}{K} \sum_{i=1}^K \ph(X_i) - Y_i \right] = \frac{1}{K} \EE_{P} \left[p(X_1) (1 - p(X_1))\right].
\end{equation}
\end{minipage}
\vspace{0pt}
\end{proposition}
}

Proposition~\ref{prop:to_size_a}a) shows that finer groups or partitions can detect more problems with the predictor.
Relating this to the above-discussed problem of stereotyping in the fairness context gives an argument for using finer groupings or partitions.
However, while larger groups may overlook more forms of stereotyping, it is also not feasible to go arbitrarily small:
Smaller groups come at the cost of giving more 'unstable' calibration errors in the sense that a group of $K$ datapoints drawn independently from a true distribution $P$ has an error variance that is inversely proportional to~$K$ (Proposition~\ref{prop:to_size_var}). 
Note that the forms of these results also highlight that the group size trade-off cannot be tackled in a general way as its character strongly depends on the distribution of data.


\subsection{Alternative grouping structures}
\label{ss:gr_structure}

We now turn to the third choice involved in the grouping of datapoints, its general structure. 
As seen e.g. in the ECE (\ref{eq:ece}), a popular way to construct groups based on a metric is by using bins.
The necessary choice of number and size of bins is further complicated by the question of whether to make the individual bins equally large or to make them contain equally many datapoints (or some other scheme).\footnote{Some considerations in the case of prediction-based bins are discussed in \citep{Nixon2019,Vaicenavicius2019}.}
Still, bins are but one specific example of a partition and yet another question is whether to use a partition in the first place.
They have the advantage of allowing a natural hierarchy of groupings since unions of non-intersecting groups are themselves sensible groups.
One can also construct partitions based on input features, e.g. using a multi-dimensional grid instead of one-dimensional bins.
We already discussed advantages of finer input-based partitions over partitions induced by prediction-based bins above.
In the following, we discuss two alternative grouping structures, which can be localised in input space.

One option is to choose overlapping groups around individual points $x \in \XX$ or $\ph(x) \in [0,1]$ by selecting their $k$ nearest neighbours ($k$-NN) in the dataset according to some notion of distance.
In contrast to fixed-size balls, k-NN directly controls the group size, independently of the distribution and size of the dataset. 
A downside in comparison to partitions is that some datapoints may be part of many more groups than others.
\highlight{
\begin{proposition}[Group overlap for $\XX$-based $k$-NN] \label{prop:knn_membership} \ \\
Take $\XX=\RR^d$ and the grouping $\{I_j\}_{j \in \II}$ based on $k$-NN of datapoints $x_1, ..., x_N$ using any $L^p$-based metric on $\XX$ with $p \in \NN$.
Depending on the dataset, it is possible that one datapoint is part of $\min(N,\ 2 \cdot d \cdot (k-1) + 1)$ groups while another datapoint is part of only one group (unless $N \leq k$).
\end{proposition}
}

When group errors are agglomerated via a function satisfying Translation Equivariace for the purpose of evaluating prediction quality (Section~\ref{s:scores}), this may result in an unwanted disproportionate influence of some datapoints compared to others.
However, $\XX$-based $k$-NN not only provides errors that can be localised in input space, it can even be shown that its finite data estimate of the calibration errors converge to their 'true' value in the limit of infinite data (assuming the existence of a true distribution $P$) when $\XX = \RR^d$.
In other words, such groups can provide consistent estimators of the true individual calibration error $\ph(x) - \EE_P[Y | X=x]$ at each $x \in \XX$.
Note that the existence of a true probability distribution is a questionable assumption.
Still, such theoretical results are always quite a positive sign.
\highlight{
\begin{proposition}[Strong consistency of $\XX$-based $k$-NN] \label{prop:knn_consist} \ \\
Assume $\XX=\RR^d$ with a metric $d_\XX$ s.t. $(\XX, d_\XX)$ is separable, a predictor $\ph: \XX \to [0,1]$, and a joint distribution $P$ on random variables $(X,Y)$ with values in $\XX \times \{0,1\}$. Denote the Bayes-optimal predictor by $p(x) := \EE[Y | X=x]$. 
Then
\begin{equation}\nonumber
    \EE_{P}\left[\left\vert c(I_X^N) - (\ph(X) {-} \ptr(X))\right\vert\right] \to 0 \text{ almost surely as } N \to \infty,
\end{equation}
where $I_x^N$ denotes the set of $k$-NN of $x$ with $k \to \infty$, $\frac{k}{N} \to 0$ for $N \to \infty$.\footnote{This slightly simplified formulation glosses over technicalities regarding the treatment of ties. For a comprehensive discussion, see \citep{Devroye1994}.
Our proof reduces the problem to $k$-NN consistency of the variable $Z := \ph(X) - Y$ (see Appendix~\ref{app:proofs}).}
\end{proposition}
}

The advantage and disadvantage of $k$-NN over partitions are also shared by kernel-based notions of calibration, to a degree determined by the chosen bandwidth.
Often based on some metric, kernels can be used to weigh the influence of datapoints in proportion to some notion of distance to $x$ or $\ph(x)$.
\citet{Luo2022} propose kernels in a learned representation space intersecting with prediction bins (see Appendix~\ref{app:established_scores}), thus interpolating between local calibration and the standard prediction binning via the bandwidth parameter.
They also show that their estimator is consistent, but with respect to the true group error rather than the individual calibration error.
To capture such kernel-based notions, we generalise calibration errors for discrete groups, defined in (\ref{eq:cal_error}), to calibration errors for normalised distributions $q$ on $\XX$:
\begin{equation}\label{eq:c_general}
    C(q) := \EE_{q}[\ph(X) - p(X)].
\end{equation}
It can easily be checked that we can express the calibration error $c(I_j)$ of a group $I_j$ through $C(q_j)$ via the distribution
\begin{equation}\label{eq:p_Ux}
    q_j(x) := \frac{|\{i \in I_j: x_i = x\}|}{|I_j|},
\end{equation}
thus generalising the notion of a group to a more abstract entity, a distribution on $\XX$.
Note that for this, we need to assume that groups always contain either all datapoints instantiating some $x \in \XX$ or none -- that is, as a union of groups $V_x$ defined in (\ref{eq:fin_freq_V}), see also footnote~\ref{fn:all_datapoints}.
There is generally no reason to include only some datapoints instantiating a given input but not others, especially since equal input also implies equal prediction.
This is corroborated by the observation that all groupings used in established calibration scores, surveyed in Appendix~\ref{app:established_scores}, satisfy this.
We can also prove the consistency of $\XX$-based kernels as estimators of the true calibration error:
\highlight{
\begin{proposition}[Strong consistency of $\XX$-based kernels] \label{prop:kernel_consist} \ \\
Assume $\XX=\RR^d$, a predictor $\ph: \XX \to [0,1]$, and a joint distribution $P$ on random variables $(X,Y)$ with values in $\XX \times \{0,1\}$. Denote the Bayes-optimal predictor by $p(x) := \EE[Y | X=x]$.
Then
\begin{equation}\nonumber
    \EE_{P}\left[\left\vert C(q_X^\gamma) - (\ph(X) {-} \ptr(X))\right\vert\right] \to 0 \text{ almost surely as } N \to \infty,
\end{equation}
where $q_x^\gamma(x') := \frac{1}{\gamma}k(x-x')\cdot \frac{1}{K}$ with $K := \sum_{i=1}^N \frac{1}{\gamma} k(x-x_i)$ denotes the normalised distribution induced by a Borel kernel $k$ with the bandwidth satisfying $\gamma \to 0$ and $\frac{N \cdot \gamma^d}{\log N} \to \infty$ for $N \to \infty$.
See Appendix~\ref{app:proofs} for the constraints on $k$.
\end{proposition}
}

Note that the bandwidth parameter brings a trade-off similar to that of the group size but allows smoother interpolation.
A general takeaway from this section should be that, as for '$\ph$ is fair', the statement '$\ph$ is calibrated' is always relative to the particular grouping.
For calibration-based evaluation, a transparent communication of the grouping choices is therefore crucial.


\FrameSep4pt
\begin{framed}
\noindent{\textbf{Intuition (Section~\ref{s:grouping}):}
    The formation of groups that is necessary for evaluating calibration involves a number of choices.
    One concerns the notion of vicinity, in particular, whether groups should be based on vicinity in input space or in prediction space (Proposition~\ref{prop:cohesion}).
    A clear trade-off concerns the size of the groups:
    smaller groups provide more detailed information (Proposition~\ref{prop:to_size_a}) while larger groups provide a more stable error estimation (Proposition~\ref{prop:to_size_var}).
    The third choice concerns the way a grouping is structured, which can be done via partitions or overlapping localised groups.
    Prime examples of the former are bins while the latter may use $k$-NN or kernels, both of which can provide consistent estimators of individual calibration scores (Propositions \ref{prop:knn_consist} and \ref{prop:kernel_consist}).}
\end{framed}


\section{Global Calibration Scores}
\label{s:scores}

Thus far, we have only considered group-wise calibration errors.
In order to evaluate predictors or to provide an objective that can be optimised, it is often helpful to condense an evaluation into a single score.
In this section, we analyse the space of sensible agglomeration functions turning group errors into a global, population-level score.
The motivation for exploring the space of sensible functions is threefold:
First, situating established agglomeration functions in this space can provide a deeper understanding of their character.
Second, it may inspire the use of alternative agglomeration functions from this space.
Third, it allows connecting the grounding of global fairness scores (Section~\ref{ss:fair_agglomeration}) in this more familiar setting, both implicitly by building intuition for agglomeration functions and explicitly by establishing mathematical connections.

In Appendix~\ref{app:established_scores}, we survey established calibration scores and show how our framework captures them, thereby allowing direct comparisons between them in terms of the choices discussed in this work. 
Two observations relating to this survey are that all the scores agglomerate \textit{absolute} calibration errors $|c(I_j)|$ \textit{via the average or the maximum}.
In line with these and all other calibration scores that we are aware of, we only consider the agglomeration of absolute calibration errors in this section, thereby taking over- and under-prediction to be equally undesirable.
In the following, we assume a set of absolute calibration errors indexed by a set $J$ on which we impose a uniform probability measure.
One could instead simply consider agglomeration functions on finite index sets, that is, on tuples of errors, but we decided to work in this more general setting that also allows working with a non-uniform measure, such as the probability $P$ on $\XX$ in case $J = \XX$.
It also allows to leverage theoretical machinery applicable to uncountable sets.
\highlight{
\begin{definition}[Agglomeration Function]\ \\
    For a measure space $J$, an \emph{agglomeration function} is a function $\Rb : \LL^2(J) \to \RR$.\label{def:agglom}
\end{definition}
}

Here, $\LL^2(J)$ is the space of real-valued random variables on the index set $J$ with finite second moment; thus, members of $\LL^2(J)$ assign to every index $j \in J$ a real number, which should be thought of as the absolute calibration error of a group $I_J$ or $q_j$.
We use the term 'agglomeration function' rather than 'aggregation function' since the latter is often defined in a more specific way \citep{Grabisch2009}.

Similar to the case of losses discussed in \citep{Williamson2019}, we suggest that for any $C, C' \in \LL^2(J)$ and any $\alpha \in \RR$, an agglomeration function $\Rb$ should satisfy the following axioms:
\ \\[3pt]
\begin{tabular}{lll}
\textbf{A1} &\textbf{Monotonicity}
&
$\Rb(C) \leq \Rb(C') \text{ if } C \leq C' \text{ almost surely}$.
\ \\
\textbf{A2} &\textbf{Translation Equivariance} 
&
$\forall \alpha \in \RR: \Rb(C + \alpha) = \Rb(C) + \alpha$.
\ \\
\textbf{A3} &\textbf{Positive Homogeneity}
&
$\Rb(0) = 0 \text{ and } \forall \mu > 0: \Rb(\mu C) = \mu \Rb(C)$.
\ \\
\textbf{A4} &\textbf{Subadditivity}
&
$\Rb(C + C') \leq \Rb(C) + \Rb(C')$.
\ \\
\textbf{A5} &\textbf{Law Invariance}
&
$\Rb(C) = \Rb(C')$ if $C$ has the same value distribution as $C'$.
\end{tabular}
\ \\[3pt]
\textbf{A1} is the requirement that larger errors should lead to larger global scores.
\textbf{A2} and \textbf{A3} are desirable properties for agglomerating group-wise calibration errors into global calibration scores as they preserve the naturally interpretable scale of the group-wise errors.
While being less interpretable, scores that do not satisfy these axioms might still be useful as objective functions.
\textbf{A4} is a particularly intuitive requirement when $C$ and $C'$ are the restrictions of some $D \in \LL^2(J)$ to some partition $\{J_0, J_1\}$: The sum of the scores on $J_0$ and $J_1$ should not be smaller than the score of $D$ (see also Appendix~\ref{app:partitions}).
Also note that A4 is equivalent to Convexity\footnote{$\Rb$ is said to be \emph{convex} if $\forall \lambda \in (0,1): \Rb((1-\lambda)C + \lambda C') \leq (1-\lambda) \Rb(C) + \lambda \Rb(C')$.\label{fn:convexity}} under A3 \citep{Artzner1999} which makes $\Rb$ an easier optimisation target.
\textbf{A5} ensures that all groups are treated equally and is sometimes also called Anonymity. Note that we may still weight groups according to their size by choosing a grouping that includes groups multiple times, as shown for the ECE in Appendix~\ref{app:established_scores} (an alternative is discussed in Appendix~\ref{app:partitions}).
It turns out that these axioms are central to the theory of risk measures, which allows to draw on insights from this rich literature.
\highlight{
\begin{definition}[Coherent Risk Measure; \cite{Artzner1999}]\ \\
    An agglomeration function satisfying A1-A4 is called \emph{coherent risk measure (CRM)}.\label{def:CRM}
\end{definition}
}

\highlight{
\begin{definition}[Conditional Value at Risk; \cite{Artzner1999}]\ \\
    Let $\qC(\alpha) := \sup \{\mu \geq 0 : P(C \geq \mu) < \alpha\}$ denote the quantile function for $C \in \LL^2(J)$, $\alpha \in [0,1]$.
    The \emph{conditional value at risk} is defined as
    $\Cvara(C) := \EE[C\ |\ C > \qC(\alpha)]$
    for $\alpha \in [0,1]$.\footnote{This definition of $\Cvara$ shows nicely that it measures the expectation of the largest $\alpha\%$ of errors but requires that $C$ has a continuous distribution which is not satisfied in practice, with finite data and groupings. Such a setting requires an alternative definition as $\Cvara(C) = \frac{1}{1-\alpha} \int_\alpha^1 \qC(t) dt$. A subclass of law-invariant CRMs that subsumes CVaR and is intuitive enough to have been independently 'discovered' by multiple disciplines are the Supremum Risk Measures \citep{Acerbi2002, Frohlich2022}.}\label{def:CVaR}
\end{definition}
}

Any law-invariant CRM $\Rb$ -- and thus any agglomeration function satisfying A1-A5 -- can be built from $\Cvara$ via
\begin{equation}\label{eq:CRMs_Kusuoka}
    \Rb(C)  = \sup_{\nu \in \Lambda}\  \int_0^1 \Cvara (C)\ d\nu(\alpha)
\end{equation}
for some set $\Lambda$ of probability measures over $[0,1]$ \citep{Kusuoka2001}.
As $\{\Cvara\}_{\alpha \in [0,1]}$ interpolates between the expectation and the maximum (attained for $\alpha = 0$ and $1$, respectively), law-invariant CRMs as a whole populate the space between expectation and maximum.
In this sense, all previously proposed calibration scores are extreme cases of sensible calibration scores.
For global scores based on $\Cvara$, the choice of $\alpha$ (between 0 and 1) determines how much attention is paid to the highest errors, or, to what extent the lowest errors are ignored.

As we have divided the design of global calibration errors into grouping and agglomeration, it is interesting to ask whether one can be reduced to the other, that is, whether choices for one of them can be 'simulated' by choices of the other.
First, note that they take into account different kinds of information.
The grouping looks at inputs and predictions without considering the resulting group errors while the agglomeration only looks at the resulting group errors.
An example where the grouping already determines the global score is when all groups have the same error: Any agglomeration function satisfying A2 \& A3 will then return the same score.
Conversely, the choice of agglomeration can also not be simulated by sensible groupings, as the maximum and the expectation generally behave very differently.
These observations suggest that the proposed organisation into grouping and agglomeration is not further reducible.

%


\FrameSep4pt
\begin{framed}
\noindent{\textbf{Intuition (Section~\ref{s:scores}):}
    By suggesting axioms that functions for agglomerating group calibration errors should satisfy, we characterise them as the class of law-invariant coherent risk measures.
    These can be seen as interpolating between the average and the maximum, which seem to be the only previously suggested agglomeration functions.}
\end{framed}


\section{Calibration-based Notions of Fairness}
\label{s:fairness}

Having laid out a general framework for designing calibration scores which assess the overall \textit{quality} of probabilistic predictions, we now turn to questions of fairness.
%
%
%
Formal scholarship on algorithmic fairness has largely focused on two types of approaches: individual and group fairness.
The demand that similar individuals be treated similarly has been proposed as a criterion for \textbf{individual fairness} \citep{Dwork2012}.
While similar treatment seems to be interpretable as similar predictions, the first 'similar' does all the work here while remaining entirely opaque.
An analogous observation has also led some legal scholars to argue that the classic formulation of such a principle in the Western tradition, Aristotle's assertion that like cases should be treated alike, is empty \citep{Westen1982}.
Another problem with such a notion of individual fairness is that it does not take into account the available data in a principled way -- it does not use it at all unless the similarity metric is constructed from it.
\textbf{Group fairness} usually describes the comparison of summary statistics between selected groups, e.g. based on race or gender.
One problem with this is that group-level statistics have little to say about the fair treatment of individuals and that they can overlook discrimination across division lines that are unaccounted for. 
In particular, it reduces individuals to their membership in one particular group that the individual is thought to belong to.
One such group-level statistic, proposed in \citep{Kleinberg2016}, is calibration, usually evaluated using pre-determined prediction space bins.
More recently, scholars have started to bridge the gap from group to individual fairness by considering multitudes of potentially overlapping groups simultaneously, under the name of subgroup fairness \citep{Kearns2018} or \textbf{multi-calibration} \citep{Hebert-Johnson2018}.
An open question is the formation of such groups, with the suggestions including all easily computable groups \citep{Hebert-Johnson2018} or groupings based on similarity of learned representations \citep{Burhanpurkar2021}.
Any such grouping is usually subdivided through some prediction-based binning; this is slightly generalised in \citep{Gopalan2022}.

In this section, we connect the richness of calibration laid out in previous sections to the debate on algorithmic fairness.
We draw on our characterisation of agglomeration functions for scores of prediction \textit{quality} in Section~\ref{s:scores} to propose and axiomatise agglomeration functions for global \textit{fairness} scores in Section~\ref{ss:fair_agglomeration}. 
In Section~\ref{ss:fair_notions}, we demonstrate the applicability of global calibration scores based on partitions and localised groups to notions of group and individual fairness, respectively.


\subsection{Global fairness scores}
\label{ss:fair_agglomeration}

Previous work on calibration for fairness has tended to not discuss global fairness scores, which would allow a direct comparison between algorithms as well as provide an objective to optimise.
The original papers \citep{Kleinberg2016, Chouldechova2017} only considered bin-wise errors without agglomerating them, while the algorithm proposed in \citep{Hebert-Johnson2018} runs until the absolute calibration error on every group-bin-combination is below a fixed value, thus implicitly enforcing a constraint on the maximum of all absolute errors.
Global fairness scores would need to take into account the signed errors, as we often care much more about whether $\ph$ over- or under-predicts:
When $c(G_1) = \mu$ and $c(G_2) = -\mu$ for some $\mu > 0$, the absolute errors of groups $G_1$ and $G_2$ are equal but this is not necessarily fair: 
When under-prediction of risk is positive from the individuals' perspective, over-prediction is often negative and vice versa.
While not every application may have this aspect, it should be possible to distinguish between under- and over-prediction.
Furthermore, while adding some constant to all errors changes the quality of the predictor, it does  not affect the inequality between the groups.

Caring not about the average calibration error of the groups but about the disparity between them requires measures of deviation rather than risk.\footnote{A strand of research that is in some ways similar is that of inequality measures, which have also been proposed for evaluating inequality of losses \citep{Speicher2018}. In our setting where calibration errors can be positive or negative and have a fixed scale, inequality measures seem less suitable. \citet{Williamson2019} propose deviation measures on losses for capturing fairness and relate them to inequality measures. In settings when we need not distinguish under- from over-prediction, their framework is also applicable to calibration.}
However, as we show in Proposition~\ref{prop:quadrangle}, there is a close connection between the two.
Furthermore, as explained above, we now take 'raw' signed calibration errors as inputs rather than their absolute values.
In addition to A3-A5, we suggest that agglomeration functions $\Db$ for global fairness scores should satisfy the following axiom for $C \in \LL^2(J)$:
\ \\[3pt]
\begin{tabular}{lll}
\textbf{A6} &\textbf{Normalisation}\quad\quad
 & $\Db(C) \geq 0$ and $\Db(C) = 0 \Leftrightarrow C$ is constant.
\end{tabular}
\ \\[3pt]
Normalisation ensures that the score is positive if and only if there is some inequality in terms of calibration errors.
A3-A5 are again desirable for the reasons discussed in Section~\ref{s:scores}.
We now introduce a new class of functions.
\highlight{
\begin{definition}[Fairness Deviation Measures]\ \\
    A \emph{fairness deviation measure (FDM)} is an agglomeration function satisfying A3-A6.\label{def:FDM}
\end{definition}
}

We can gain some insights into this class by exploiting a handy connection to the law-invariant coherent risk measures discussed in the previous section. 
This connection is characterised by the following result which is basically a corollary of the Quadrangle Theorem in \citep{Rockafellar2013}.\footnote{Strictly speaking, the measures must also satisfy that for all $\alpha \in \RR$, $\{C \in \LL^2(J): \Rb(C) \leq \alpha\} \text{ is closed}$. This is included as a further axiom in \citep{Rockafellar2013} but we do not dwell on this since it is automatically satisfied for convex finite measures on finite $J$, which is always the case in practice.\label{fn:closedness}}
It allows the construction of FDMs from CRMs and vice versa, shedding light on the relationship between quality and fairness scores.
\highlight{
\begin{proposition}[Relation between CRMs and FDMs]\label{prop:quadrangle} \
    \begin{enumerate}[a)]
    \item[a)] Agglomeration functions $\Db$ satisfying A3-A6 stand in a one-to-one relationship with agglomeration functions $\Rb$ satisfying A2-A5 and Aversity (the property that $\Rb(C) > \EE[C]$ if $C \in \LL^2(J)$ is  not constant) via 
    \begin{minipage}{\linewidth}
    \vspace{4pt}
    \begin{equation}\label{eq:quadrangle}
        \Rb(C) = \EE[C] + \Db(C).
    \end{equation}
    \end{minipage}
    \vspace{-0pt}
    \item[b)] A law-invariant CRM induces an FDM via (\ref{eq:quadrangle}) iff it is averse.
    \item[c)] An FDM induces a law-invariant CRM via (\ref{eq:quadrangle}) iff it satisfies $\Db(C) \leq \sup C - \EE[C]\ \forall C \in \LL^2(J)$.
    \end{enumerate}
\end{proposition}
}
%
\highlight{
\begin{example}[Fairness Deviation Measures]
    $$\setcounter{footnote}{15}
    \begin{array}{lll}
        \hbox{1. Superquantile deviation:} & \Db(C) = \Cvara(C - \EE[C])\ \text{for }  \alpha > 0 
        &\quad\quad\quad\quad\quad\quad\quad\quad\quad\quad\quad\quad\quad\quad\quad\\
        \hbox{2. Standard deviation:} & \Db(C) = \sigma(C)
        &\quad\quad\quad\quad\quad\quad\quad\quad\quad\quad\quad\quad\quad\quad\quad\\
        \hbox{3. Range deviation$^{16}$:} & \Db(C) = \sup C - \inf C
        &\quad\quad\quad\quad\quad\quad\quad\quad\quad\quad\quad\quad\quad\quad\quad
    \end{array}$$
\end{example}
}

The\setcounter{footnote}{16}\footnotetext{This is not an unfamiliar score for other fairness metrics: \citet{wang2022} discuss it under the more intuitive notion of a 'maximum difference [...] of a performance metric [...] between two groups'.} superquantile deviation is induced by $\Cvara$ via (\ref{eq:quadrangle}).
Both standard deviation and range deviation do not induce CRMs as they do not satisfy $\Db(C) \leq \sup C - \EE[C]\ \forall C \in \LL^2(J)$.
More examples of risk-deviation-pairs can be found in \citep{Rockafellar2013}.
Recall, however, that the coherent risk measures in the previous section are assumed to take absolute calibration errors as inputs while the fairness deviation measures take the 'raw' signed errors.


\subsection{Group fairness and individual fairness}
\label{ss:fair_notions}

In the remainder of this section, we show that global calibration scores based on different choices of grouping structure (Section~\ref{ss:gr_structure}) can reflect different notions of fairness.
As mentioned above, previous work on calibration and fairness did not provide such population-level scores but largely aimed to compare or minimise absolute group-level errors.
The use of global scores based on deviation measures (Section~\ref{ss:fair_agglomeration}) allows a more comprehensive assessment of fairness than previous approaches which compare bin-wise calibration errors between groups of interest.
In cases where comparisons between prespecified, non-overlapping groups are desired, partitions may be the most suitable structure for evaluating calibration as they can provide strict boundaries along such group lines.
Calibration can, thus, provide a notion \textbf{(sub-)group fairness} when based on some form of partition; this observation is also in accordance with much of the previous literature on calibration-based fairness.

In Sections \ref{s:local_cal} and \ref{ss:gr_vicinity}, we pointed to the availability of alternatives to prediction-based grouping.
While prediction-based bins for calibration have been suggested as a way to assess whether probabilities in different groups 'mean the same thing' \citep{Kleinberg2016}, they are at odds with finer partitions based on input features due to constraints on the group sizes (Section~\ref{ss:gr_size}).
Finer input-based groups, however, allow to detect more forms of stereotyping and should, thus, be considered at least as a complementary basis for sub-group fairness evaluations.
A downside of prespecified groups forming a partition is that each individual is only considered as a member of one particular (possibly intersectional) group, which is, after all, a form of pigeonholing.
In what follows, we argue that calibration also allows for notions of individual fairness.

A problem with previous (Lipschitz-based) notions of \textbf{individual fairness} is that they do not require what \citet{Vredenburgh2022}, following \citep{Broome1990}, takes to be the essence of fairness, namely 'respecting people’s claims in proportion to the strength of those claims’.
Calibration offers a principled way to assess the strength of claims by automatically comparing probabilistic predictions with the available data.
Failure to take into account data is also criticised by \citet[p. 60]{Selbst2019} who argue that '[w]ithin the algorithmic frame, any notion of ''fairness'' cannot even be defined'.
If we assumed the existence of and access to a true probability distribution $P(Y|X)$ (both of which are problematic), this would provide a measure for the strength of people’s claims without reference to data:
One could see the true probability (e.g. of credit default) to provide grounds for a proportional claim to a particular treatment (e.g. prioritisation in granting loans).
Equality of the true calibration error $\ph(x) - \EE_P[Y | X=x]$ across the population would ensure that the strengths of these claims are respected (although other justice-related claims may not be).
However, the theoretical notion of the true calibration error cannot provide a notion of individual fairness in practice as we never have access to the purported true probability distribution.
We now argue that in practice, calibration on localised groups can provide a notion of individual fairness by grounding claims in available data -- if the assumption of representative data is justified.

As shown in Section~\ref{s:local_cal}, a 'natural' way to construct groups for calibration is as extensions \textit{around} an individual $x$, in order to make up for limited data \textit{at} $x$.
Furthermore, as established in Propositions~\ref{prop:knn_consist} and~\ref{prop:kernel_consist}, we can get consistent estimators for the individual calibration error at given points in input space through suitable choices of grouping.
Calibration errors of such cohesive groups (or kernels) that can be localised to a particular point in input space and take into account the surrounding datapoints, hence, provide a practical and quantifiable notion of respecting people's claims that is also theoretically sound.
A potential downside of such a notion is that for localised groupings, some datapoints can be members of multiple or even all groups (Proposition~\ref{prop:knn_membership}) and may, thus, be thought to have a strong influence on the global fairness score.
Note, however, that this is not the case for the measures introduced above that assess the deviation between group-wise calibration errors, quite on the contrary:
For a deviation measure like the standard deviation, membership in all groups implies that a datapoint has no influence at all on the global score.
In sum, global calibration scores for suitable localised groups (e.g. based on $k$-NN or kernels) can allow to assess whether individuals receive risk scores in proportion to the data available for them which, we posit, is a notion of individual fairness -- this is also in line with \citet[p. 199]{Broome1991} noting that his notion of unfairness, failure to proportionally satisfy claims, 'is plainly an individual harm’.

\FrameSep4pt
\begin{framed}
\noindent{\textbf{Intuition (Section~\ref{s:fairness}):}
    Previous literature has discussed different notions of fairness, notably group and individual fairness.
    Although calibration is seen as a promising approach to (sub-)group fairness, there is no established metric for it.
    Potential calibration-based global fairness scores need to take into account the \textit{signed} errors and focus more on their \textit{disparity}.
    While this implies that they are quite different to the scores discussed in the previous section, they are closely connected.
    Depending on the grouping structure, global calibration-based fairness scores using deviation measures can provide notions of individual or subgroup fairness.}
\end{framed}


\section{Conclusion and Future Work}

By tracing its roots and investigating implicit choices, we have provided a general framework that captures the richness of calibration from the ground up.
We placed particular emphasis on groupings based on the input space and motivated the consideration of groups that can be localised there.
We showed that different ways of evaluating calibration correspond to different notions of fairness. 
In the same way as assertions about fairness always need to specify the 'with regard to what?', calibration depends on choices of grouping.
For a more solid understanding of this relationship, further formal and conceptual investigations will be necessary.
While we made some first steps to analyse the trade-offs inherent in choices of grouping, we see much potential for more comprehensive results.
Likewise, while we show a connection between the agglomeration functions suitable for evaluating prediction quality and for evaluating fairness, further work may allow to characterise (im)possibilities of their joint optimisation. 
To mitigate the dependence on the chosen grouping, it may help to instead use multiple groupings, such as a sequence of increasingly fine partitions; frameworks for utilising such partially redundant information also seem a worthwhile endeavour.
%
%
%
In general, we envision our work as a fertile ground for further research on calibration and its role in fairness in particular.
We conclude by again stressing that calibration can only be one element in the just deployment of algorithms in society, given that 'accurate modelling within unjust institutions can replicate and further injustice'
\citep{Vredenburgh2022}.



\begin{acks}
This work was funded by the Deutsche Forschungsgemeinschaft (DFG, German Research Foundation) under Germany’s Excellence Strategy—EXC number 2064/1—Project number 390727645 as well as the German Federal Ministry of Education and Research (BMBF): Tübingen AI Center, FKZ: 01IS18039A. 
The authors thank the International Max Planck Research School for Intelligent Systems (IMPRS-IS) for supporting Benedikt Höltgen.
Big thanks also to Rabanus Derr, Christian Fröhlich, and the anonymous reviewers for thoughtful feedback.
\end{acks}

\bibliographystyle{ACM-Reference-Format}
\bibliography{library}

\newpage
\appendix


\section{Established Calibration Scores}
\label{app:established_scores}

Recall our setting of datapoints $\{(x_i, y_i)\}_{i \in \II}$ with $\II = \{1,...,N\}$.
In (\ref{eq:cal_error}), we defined the calibration error of a group $I \subset \II$ as
\begin{equation}\label{eq:c}
    c(I) := \frac{1}{|I|} \sum_{i \in I} \ph(x_i) - y_i.
\end{equation}
We now demonstrate how different notions of calibration can be seen as some form of grouping combined with some form of agglomeration of errors.
For this, it is useful to think in terms of a parametrised set of groups $\{I_j\}_{j \in J}$ for some index set $J$ which can take different forms.
As a first example, we can write the ECE described in (\ref{eq:ece}) for bins $B_1, ..., B_K$ of $[0,1]$ as
\begin{equation}\label{eq:aggr_ex_ece}
    \ECE(\ph) = \frac{1}{N} \sum_{j = 1}^N |c(I_j)|, \quad I_j := \{i \in \II : \ph(x_i) \in B_{k(j)}\},
\end{equation}
with $B_{k(j)}$ denoting the bin in which $\ph(x_j)$ falls.
Note that the weighting by the size of the bins is now done implicitly by adding one term for each datapoint, rather than one term per bin.
We could avoid this implicit weighting by choosing a different grouping which nevertheless includes the same groups, just in different multiplicities.
A score that weights each bin equally, we may call it average calibration error (ACE), is given by 
\begin{equation}\label{eq:aggr_ex_mce}
    \ACE(\ph) = \frac{1}{K} \sum_{k = 1}^K |c(I_k)|, \quad I_k := \{i \in \II: \ph(x_i) \in B_k\}.
\end{equation}
Using the same groups $I_k$, we can also define the sometimes invoked maximum calibration error (MCE; \cite{Naeini2015}) in this framework as 
\begin{equation}
    \MCE(\ph) = \max_k |c(I_k)|.
\end{equation}

Interestingly, the well-known Brier score is also equivalent to a function over calibration errors for a suitable grouping.
As shown in \citep{Sanders1963} for a finite prediction space, the Brier score can be decomposed into two terms corresponding to notions of calibration and refinement/sharpness, respectively:
\begin{equation}\label{eq:BS_y}
    \frac{1}{N} \sum_{i=1}^N [\ph(x_i) - y_i]^2
     = \sum_{v \in \YY_D} \Ph(v) [\ptil(v) - v]^2  +  \sum_{v \in \YY_D} \Ph(v) \ptil(v) (1 - \ptil(v))
\end{equation}
where $\YY_D := \{\ph(x_i) : i \in \II \}$ is the set of predictions on the observed datapoints, $\Ph(X, Y, \ph(X))$ denotes the empirical distribution on $\XX \times \{0,1\} \times [0,1]$, and $\ptil(v) := \Ph(Y=1 | \ph(X)=v)$ denotes the observed ratio of positive labels for datapoints where $\ph$ predicts $v$.
This can be seen as grouping by level sets of $\ph$.
Instead grouping the datapoints by inputs allows the decomposition
\begin{align}
\frac{1}{N} \sum_{i=1}^N\ (\ph(x_i) - y_i)^2 \label{eq:deriv_start}
    &= \sum_{x \in \XX_\DD} P(x) \left[(\ph(x) - 1)^2 \ptr(x) + (\ph(x) - 0)^2 (1 - \ptr(x))\right] \\
    &= \sum_{x \in \XX_\DD} P(x) \left[\ph(x)^2 \ptr(x) {-} 2 \ph(x) \ptr(x)  {+} \ptr(x) + \ph(x)^2 - \ph(x)^2 \ptr(x)\right]\\
    &= \sum_{x \in \XX_\DD} P(x) \left[\ph(x)^2 - 2 \ph(x) \ptr(x)  + \ptr(x)^2 - \ptr(x)^2 + \ptr(x) \right]\\
    &= \sum_{x \in \XX_\DD} P(x) [\ph(x) - \ptr(x)]^2 + \sum_{x \in \XX_\DD} P(x) \ptr(x) [1 - \ptr(x)] \label{eq:deriv_end}
\end{align}
with $P(X,Y)$ denoting the empirical distribution on $\XX \times \{0,1\}$ and $p(x) := \EE_P(Y | X=x)$.
Note the second term of (\ref{eq:deriv_end}) is independent of the predictor $\ph$, implying that minimising the Brier score is equivalent to minimising the first term.\footnote{This tells us that the Brier score cannot reduce to zero even for the Bayes optimal predictor $\ph = p$, unless $\forall x \in \XX : p(x) \in \{0,1\}$.
This makes sense since the terms $(\ph(x_i) - y_i)^2$ can only be all zero if no $x$ is observed with both labels, i.e. $\EE_P(Y | X=x)$ only takes values in $\{0,1\}$.
A comparison with (\ref{eq:single_var}) reveals that the second term is the calibration error variance of a point drawn randomly from the empirical distribution!}
This first term can be seen as a function of calibration errors:
\begin{equation}\label{eq:aggr_ind_brier}
    \frac{1}{N} \sum_{i = 1}^N c(I_i)^2, \quad I_i := \{i \in \II : x_i = x\}.
\end{equation}
The average-of-squares does not satisfy A2 and A3 (Section~\ref{s:scores}), which may have contributed to the observation that 'still there is no clear consensus on how to interpret the Brier score' \citep[p. 85]{Cabitza2022}.

As suggested in Section~\ref{ss:gr_structure}, we can capture notions of calibration that invoke kernels through the generalised calibration error defined for distributions $q$ on $\XX$ as 
\begin{equation}
    C(q) := E_{q}[\ph(X) - p(X)].
\end{equation} 
Kernels have been proposed on the prediction space \citep{Kumar2018} and on a learned representation space intersected with bins in the prediction space \citep{Luo2022}.
The maximum local calibration error (MLCE) proposed in \citep{Luo2022}, which we briefly discussed in Section~\ref{ss:gr_structure}, can in our framework be defined as
\begin{equation}\label{eq:MCLE}
     \MLCE = \max_{x \in \XX_\DD} C(q_x)
\end{equation}
for
\begin{equation}
     q_x(x') \propto k_\gamma(x,x') \cdot \mathbf{1}_{\ph(x') \in B_{k(x)}}(x'),
\end{equation}
where $k_\gamma$ is some kernel on a learned representation space with bandwidth $\gamma$ and $B_{k(x)}$ denoting the bin (within a pre-defined partition of $[0,1]$) into which $\ph(x)$ falls.
Their local calibration error (LCE) is simply the generalised group calibration error $C(q_x)$.



\section{On Group Cohesion and Topology}
\label{app:topology}

While criteria of group selection have received some attention within Fair-ML in recent years, they have largely ignored the property of group cohesion -- probably partly because the salient groups are cohesive by default:
Groups defined by a particular value or interval of a feature such as gender, race or age -- or an intersection thereof -- are intuitively cohesive;
they are even convex, as any point that, in input space, lies between two points belonging to some such group will also belong to that same group.
Cohesion (in an intuitive sense) is even weaker than that as it only requires that a group's region in input space is not disconnected.
Another reason why cohesion is usually not discussed may be that it is to be defined in terms of the mathematical input space rather than the datapoints themselves, thus requiring a further abstraction.
In the context of calibration, however, groups of interest are often not defined by hand but, for example, through approximate level sets of the predictor (that is, bins); this implies that group cohesion cannot be taken for granted anymore, as shown in Proposition~\ref{prop:cohesion}.
However, cohesion seems to be a necessary condition for a group to be interpretable and perhaps to be considered ethically relevant -- although cohesion is not sufficient as, in multiple dimensions, one can cook up all sorts of gerrymandered cohesive groups.

In the present paper, we aim to highlight the topic of group coherence -- also beyond the fairness context and particularly relating to prediction-based groups -- and motivate further discussion.
With the qualification that too much abstraction can sometimes distract from concrete ethical problems, it seems that topology would be the natural mathematical language for this.
Topology is the study of geometric properties such as continuity and connectedness in most general terms.
The basic mathematical object is a topological space, which is a set equipped with a particular set structure.
The choice of set structure, the topology, is usually fairly straightforward in $\RR^d$ (see Example~\ref{ex:standard_top}) but less so when categorical features are involved.
While there are some obvious approaches using product topologies or embeddings in continuous dimensions, such questions are left open in this work.
In the following, we introduce the basic notions needed for Proposition~\ref{prop:cohesion}.

\highlight{
\begin{definition}[Topology, Open set, Neighbourhood]\ \\
    A \emph{topology} on a set $\XX$ is a collection of subsets $\Tcal \subset 2^{\XX}$ satisfying
    \begin{itemize}
        \item $\XX, \emptyset \in \Tcal$.
        \item Any union of elements in $\Tcal$ is also in $\Tcal$.
        \item Any finite intersection of elements in $\Tcal$ is also in $\Tcal$.
    \end{itemize}
    The tuple $(\XX, \Tcal)$ (or sometimes just $\XX$) is then called a \emph{topological space}.
    The elements of $\Tcal$ (which are sets in $\XX$) are called \emph{open sets}.
    A set that includes an open set containing $x \in \XX$ is called a \emph{neighbourhood} of $x$.
\end{definition}
}

\highlight{
\begin{example}[Trivial topology]\ \\
    For any set $\XX$, the \emph{trivial topology} is given by $\{\XX, \emptyset\}$.
\end{example}
}

\highlight{
\begin{example}[Induced topology, Standard topology on $\RR^d$]\label{ex:standard_top} \ \\
    For any space $\XX$ equipped with a metric $d$, the topology \emph{induced} by $d$ is the topology generated by all balls $B_\epsilon(x)$ with $x \in \XX$, $\epsilon >0$.
    That is, it contains all sets generated by unions and finite intersections of such balls as well as $\emptyset$ and $\XX$.
    In the case of $\XX = \RR^d$, all $L^p$-based metrics induce the same topology which is called the \textit{standard topology}.
    For $\XX = \RR$ in particular, open sets in this topology are exactly the open intervals plus their unions and $\emptyset$.
\end{example}
}

\highlight{
\begin{definition}[Hausdorff space]\ \\
    A topological space $(\XX, \Tcal)$ is called a \emph{Hausdorff space} if any two points have non-intersecting neighbourhoods.
\end{definition}
}

\noindent 
Any metric space is Hausdorff under the induced topology while any space with at least two points is not Hausdorff when equipped with the trivial topology.

\highlight{
\begin{definition}[Topological continuity]\ \\
    A function between two topological spaces is \emph{continuous} if the preimage of each open set is an open set.
\end{definition}
}

\noindent 
This topological notion of continuity coincides with the better-known $\epsilon$-$\delta$-notion of continuity when the topologies are induced by metrics.
Another important notion in general topology and also in the present work is that of connectedness.
\highlight{
\begin{definition}[Connectedness]\ \\
    A set in a topological space is \emph{connected} if it is not the union of two disjoint nonempty open sets.
\end{definition}
}

\noindent 
A perhaps more intuitive notion of cohesion is path-connectedness.

\highlight{
\begin{definition}[Path-connectedness]\ \\
    A \emph{path} from a point $x$ to a point $x'$ in a topological space $\XX$ is a continuous function $\gamma : [0,1] \to \XX$ with $\gamma(0) = x$ and $\gamma(1) = x'$.
    A set in a topological space is \emph{path-connected} if for any two points in it there is a path between them.
\end{definition}
}

\noindent 
Path-connectedness is weaker than connectedness in the sense that every path-connected set is connected but not every connected set is path-connected.
\vspace{10pt}


\section{Proofs}
\label{app:proofs}




\textbf{Proposition \ref{prop:cohesion} (Cohesion in domain and codomain).} \ 
\begin{enumerate}[a)]
    \item[a)] For any continuous function $f: \XX \to \ZZ$, connectedness in $\XX$ implies connectedness in $\ZZ$ but not vice versa.
    \item[b)] For a non-constant predictor $\ph : \XX \to [0,1]$ with the standard topology on $[0,1]$, there is a topology on $\XX$ s.t. $\ph$ is not continuous (and thus connectedness in $\XX$ does not imply connectedness in $[0,1]$).
    \item[c)] If $\ph: \XX \to [0,1]$ has more than a single local minimum or more than a single local maximum, there exist intervals in $[0,1]$ whose preimages in $\XX$ are not connected (in any Hausdorff topology on $\XX$).
\end{enumerate}
\begin{proof}\
\begin{enumerate}[a)]
    \item[a)] 
    It is a standard result in general topology that the image of a connected set under a continuous map is also connected.
    A counterexample to the inverse direction is $\XX = [-1,1]$,  $\ZZ = [0,1]$ and $f: x \mapsto x^2$ with the standard topologies on $[-1,1]$ and $[0,1]$: the preimage of the connected set $[\frac{1}{4}, 1]$ is $[-1, -\frac{1}{2}] \cup [\frac{1}{2}, 1]$ which is not connected.
    \item[b)]
    Take the trivial topology $\Tcal = \{\emptyset, \XX\}$ on $\XX$, take two distinct values $v_1, v_2 \in [0,1]$ that $\ph$ realises on $\XX$. 
    Then take any open interval $I$ around $v_1$ which does not contain $v_2$ (which is possible since $v_1 \neq v_2$ and $[0,1]$ with the standard topology is Hausdorff). 
    Its preimage $\ph^{-1}[I]$ is neither $\emptyset$ nor $\XX$, so not open in the trivial topology. Hence, $\ph$ is not continuous.
    \item[c)]
    Assume $\ph$ has two (distinct) local maxima at $x_a$ and $x_b$, with $\ph(x_a) \leq \ph(x_b)$ (the proof for two minima is analogous).
    We operationalise the notion of a local maximum $x_a$ in a Hausdorff space as the property that any connected, non-singleton neighbourhood of $x_a$ contains a point $x$ such that $\ph(x) < \ph(x_a)$.
    Now consider the interval $I:=[\ph(x_a), \ph(x_b)] \subset [0,1]$ under $\ph$.
    If its preimage were connected, there would be a point $x \in \ph^{-1}[I]$ s.t. $\ph(x) < \ph(x_a)$ since it contains the local maximum $x_a$ (which is distinct from $x_b \in p^{-1}[I]$ by assumption).
    Being the preimage of $I$, this would mean that $x \in [x_a, x_b]$ and thus $\ph(x) \geq \ph(x_a)$ which is a contradiction.
    Therefore, $p^{-1}[I]$ cannot be connected.
\end{enumerate}
\end{proof}

\ \\[-3pt]
\textbf{Proposition \ref{prop:to_size_a} (Advantage of smaller groups: Resolution).}  \ \\
Take two nonempty, disjoint sets of indices $I_1, I_2 \subset \II = \{1,...,N\}$ and recall the definition of the calibration error in (\ref{eq:cal_error}).
Then 
\begin{enumerate}[a)]
    \item[a)] Assume that $\forall (i,j) \in I_1 \times I_2 : x_i \neq x_j$ and that not all labels $y_i, i \in I_1 \cup I_2$ are equal.
    Then a predictor $\ph: \XX \to [0,1]$ can be perfectly calibrated on $I_1 \cup I_2$ but neither on $I_1$ nor on $I_2$.
    \item[b)] For any predictor $\ph: \XX \to [0,1]$, $|c(I_1)| \leq \alpha$ and $|c(I_2)| \leq \beta$ entails $|c(I_1 \cup I_2)| \leq \frac{\alpha \cdot |I_1| + \beta \cdot |I_2|}{|I_1| + |I_2|}$.
    For $\alpha = \beta$, this means $|c(I_1 \cup I_2)| \leq \alpha$.
\end{enumerate}
\begin{proof}\
\begin{enumerate}[a)]
    \item[a)]
    For $j \in \{1,2\}$, let $z_j := \frac{1}{|I_j|} \sum_{i \in I_j} y_i$ be the average label in $I_j$.
    Note that this implies 
    \begin{equation}
        \sum_{i \in I_j} z_j - y_i = 0. \label{eq:ast}
    \end{equation}
    If $z_1 \neq z_2$, then the constant predictor defined by
    \begin{equation}
        \ph(x) := \frac{1}{|I_1|+|I_2|} \sum_{i \in I_1 \cup I_2} y_i
    \end{equation}
    is perfectly calibrated on the union but not on the individual sets. So from now on now assume $z_1 = z_2$.\\
    Then by assumption, $0 < z_1 < 1$ (as otherwise, all labels would be equal).\\
    So there exists $\epsilon > 0$ s.t. $0 < z_1 - \epsilon < z_1 + \epsilon < 1$.\\
    Define $\XX_j := \{x_i : i \in I_j\} \subset \XX$ for $j \in \{1,2\}$ and choose a $\ph$ such that
    \begin{equation}
    \ph(x) := 
    \begin{cases}
        z_1 + \frac{|I_2|}{|I_1|+|I_2|}\epsilon & \text{ if } x \in \XX_1 \\
        z_1 - \frac{|I_1|}{|I_1|+|I_2|}\epsilon & \text{ if } x \in \XX_2.
    \end{cases}
    \end{equation}
    Then 
    \begin{align}
        \frac{1}{|I_1 \cup I_2|} \left( \sum_{i \in I_1 \cup I_2} \ph(x_i) - y_i \right)
        &=\
        \frac{1}{|I_1| + |I_2|} \left( \sum_{i \in I_1} \left[z_1 + \frac{|I_2|}{|I_1|+|I_2|}\epsilon - y_i\right] 
        + \sum_{i \in I_2} \left[z_1 - \frac{|I_1|}{|I_1|+|I_2|}\epsilon - y_i\right] \right) \\
        &=\
        \frac{1}{|I_1| + |I_2|} \left( |I_1| \cdot \frac{|I_2|}{|I_1|+|I_2|}\epsilon - |I_2| \cdot \frac{|I_1|}{|I_1|+|I_2|}\epsilon \right) \\
        &=\ 0,
    \end{align}
    using (\ref{eq:ast}), whereas for $j \in \{1,2\}$,
    \begin{align}
        \frac{1}{|I_j|} \left\lvert \sum_{i \in I_j} \ph(x_i) - y_i \right\rvert
        =\
        \frac{1}{|I_j|} \left\lvert \pm \frac{|I_1| \cdot|I_2|}{|I_1|+|I_2|}\epsilon \right\rvert
        > 0.
    \end{align} 
    \item[b)] 
    Assume that $|c(I_1)| = \frac{1}{|I_1|} \left\lvert \sum_{i \in I_1} \ph(x_i) - y_i \right\rvert \leq \alpha$ 
    and $|c(I_2)| = \frac{1}{|I_2|} \left\lvert \sum_{i \in I_2} \ph(x_i) - y_i \right\rvert \leq \beta$.
    Then,
    \begin{align}
        |c(I_1 \cup I_2)|
        &=\
        \frac{1}{|I_1 \cup I_2|} \left\lvert \sum_{i \in I_1 \cup I_2} \ph(x_i) - y_i \right\rvert \\
        &=\
        \frac{1}{|I_1| + |I_2|} \left\lvert \sum_{i \in I_1} [\ph(x_i) - y_i]  + \sum_{i \in I_2} [\ph(x_i) - y_i] \right\rvert \\
        &\leq\
        \frac{1}{|I_1| + |I_2|} \left( \left\lvert \sum_{i \in I_1} [\ph(x_i) - y_i] \right\rvert  + \left\lvert \sum_{i \in I_2} [\ph(x_i) - y_i] \right\rvert \right) \\
        &\leq\
        \frac{\alpha \cdot |I_1| + \beta \cdot |I_2|}{|I_1| + |I_2|}.
    \end{align}
\end{enumerate}
\end{proof}

\ \\
\textbf{Proposition \ref{prop:to_size_var} (Advantage of larger groups: Variance).} \ \\
Let $(X_1, Y_1), ..., (X_K, Y_K)$ be pairs of i.i.d. random variables with values in $\XX \times \{0,1\}$ and joint distribution $P$. 
Denote the Bayes-optimal predictor by $p(x) := \EE_P[Y | X=x]$ and take some predictor $\ph: \XX \to [0,1]$. 
Then
\begin{equation}
    \var_P \left[\frac{1}{K} \sum_{i=1}^K \ph(X_i) - Y_i \right] = \frac{1}{K} \EE_{P} \left[p(X_1) (1 - p(X_1))\right].
\end{equation}
\begin{proof}\ \\
For a specified point $x \in \XX$, we get
\begin{equation}
    \EE_{P(Y | X=x)} [\ph(x) - Y] 
    = [\ph(x) - 1] \ptr(x) + \ph(x) [1 - \ptr(x)] 
    =  \ph(x) - \ptr(x)
\end{equation}
and, with the same steps as in derivation (\ref{eq:deriv_start})-(\ref{eq:deriv_end}),
\begin{equation}
    \EE_{P(Y | X=x)}[(\ph(x) - Y)^2] = [\ph(x) - \ptr(x)]^2 + \ptr(x) [1 - \ptr(x)].
\end{equation}
This gives 
\begin{equation}
    \var_{P}[\ph(X_1) - Y_1] 
    = \EE_{P}[(\ph(X_1) - Y_1)^2]  - \EE_{P}[\ph(X_1) - Y_1]^2
    = \EE_{P}[\ptr(X_1) (1 - \ptr(X_1))].\label{eq:single_var}
\end{equation}
Now since the $(X_1,Y_1), ..., (X_K, Y_K)$ are i.i.d., we can compute
\begin{equation}
    \var_{P} \left[ \frac{1}{K} \sum_{i=1}^K \ph(X_i) - Y_i \right]
    = \frac{1}{K} \var_{P}[\ph(X_1) - Y_1]
    = \frac{1}{K} \EE_{P}[\ptr(X_1) (1 - \ptr(X_1))].
\end{equation}
\end{proof}

\ \\
\textbf{Proposition \ref{prop:knn_membership} (Group overlap for $\XX$-based $k$-NN).} \ \\
Take $\XX=\RR^d$ and the grouping $\{I_j\}_{j \in \II}$ based on $k$-NN of datapoints $x_1, ..., x_N$ using any $L^p$-based metric on $\XX$ with $p \in \NN$.
Depending on the dataset, it is possible that one datapoint is part of $\min(N,\ 2 \cdot d \cdot (k-1) + 1)$ groups while another datapoint is part of only one group (unless $N \leq k$).
\begin{proof}\ \\
First assume the case $N = 2 \cdot d \cdot (k-1) + 1$.\\
We construct a dataset $x_1, ... x_N \in \RR^d$ s.t. for each dimension $s \in \{1, ..., d\}$, there are $(k-1)$ points between $\mathbf{0}$ and $\pm \mathbf{e_s}$ (with $\mathbf{e_s}$ denoting the standard unit vector of $\RR^d$ along dimension $s$), except that in dimension $1$, one point is at $3 \cdot \mathbf{e_1}$ (cf. Figure~\ref{fig:prop4}):\\
For $s \in \{1, ..., d\}$ and $(s-1) \cdot (k-1) < i \leq s\cdot(k-1)$, let $x_i := \mu \cdot \mathbf{e_s}$ (except $x_1$) and $x_{i + d\cdot(k-1)}:= - \mu \cdot \mathbf{e_s}$ for any choice or choices of $\mu \in (0,1)$.
Now let $x_1 := 3 \cdot \mathbf{e}_1$ and $x_N := \mathbf{0}$.\\
Then the point $x_N$ is among the $k$ nearest points for every point $x_i, i \in \{1,...,N\}$ (as $\|x_N - x_i\|_p = \|x_i\|_p$ ) whereas $x_1$ is among the $k$ nearest points only of itself.\\
\begin{figure}[H]
    \centering
    \includegraphics[width=.5\linewidth]{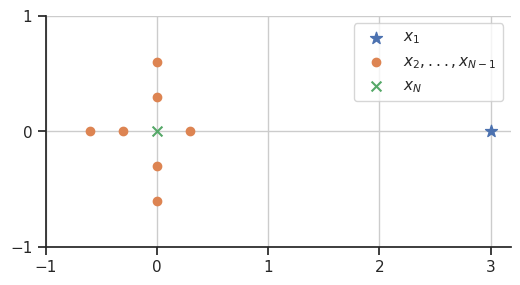}
    \caption{Illustration of the proof of Proposition~\ref{prop:knn_membership}, for $d=2$ and $k=3$, case $N = 2 \cdot d \cdot (k-1) + 1$ $(= 9)$.}
    \label{fig:prop4}
\end{figure}
\noindent
For the case $N < 2 \cdot d \cdot (k-1) +1$, we assign less than $2 \cdot d \cdot (k-1) -1$ points $x_2, ..., x_{N-1}$ between $\mathbf{0}$ and $\pm \mathbf{e_s}$, such that $x_N$ is still part of all groups while $x_1$ is still furthest from all points.\\
For the case $N > 2 \cdot d \cdot (k-1) +1$, we assign the additional points $x_{2 \cdot d\cdot(k-1) +1}, ..., x_{N-1}$ to $-2 \cdot \mathbf{e_1}$ which changes neither that $x_N$ is among the $k$-nearest points of all $x_2, ..., x_{2 \cdot d\cdot(k-1)}$, nor that $x_1$ is furthest from all points.
\end{proof}

\ \\
\textbf{Proposition \ref{prop:knn_consist} (Strong consistency of $\XX$-based $k$-NN).}  \ \\
Assume $\XX=\RR^d$ with a metric $d_\XX$ s.t. $(\XX, d_\XX)$ is separable, a predictor $\ph: \XX \to [0,1]$, and a joint distribution $P$ on random variables $(X,Y)$ with values in $\XX \times \{0,1\}$. Denote the Bayes-optimal predictor by $p(x) := \EE[Y | X=x]$.
Then
\begin{equation}
    \EE_{P}\left[\left\vert c(I_X^N) {-} (\ph(X) - \ptr(X))\right\vert\right] \to 0 \text{ almost surely as } N \to \infty,
    \nonumber
\end{equation}
where $I_x^N$ denotes the set of $k$-NN of $x$ with $k \to \infty$, $\frac{k}{N} \to 0$ for $N \to \infty$.
\begin{proof}\ \\
We can reduce our problem to the consistency of $k$-NN estimators; in the following we adapt notation used in \citep{Devroye1994}:
For fixed $\ph$, define $Z := \ph(X) - Y$.
Then $Z$ is bounded and we get a joint distribution $(X,Z)$, a true calibration error $m(x) = \EE[Z|X=x] = \ph(X) - \ptr(x)$ and an estimator $m_N(x) := c(I_x^k) = \frac{1}{k} \sum_{i \in I_x^k} z_i$ with $z_i = \ph(x_i) - y_i$.\\
Then by Theorem 1 from \citep{Devroye1994}, we get 
$\EE \left[\left\vert m_N(x) - m(x) \right\vert\right] \to 0$ with probability 1 as $N \to \infty$, which is our desired result.\footnote{Note that we are glossing over the details of tie-breaking, as the expectation is also an expectation over a supporting variable $Z$.}
\end{proof}

\ \\
\textbf{Proposition \ref{prop:kernel_consist} (Strong consistency of $\XX$-based kernels).} \ \\
Assume $\XX=\RR^d$, a predictor $\ph: \XX \to [0,1]$, and a joint distribution $P$ on random variables $(X,Y)$ with values in $\XX \times \{0,1\}$. Denote the Bayes-optimal predictor by $p(x) := \EE[Y | X=x]$.
Then
\begin{equation}
    \EE_{P}\left[\left\vert C(q_X^\gamma) {-} (\ph(X) - \ptr(X))\right\vert\right] \to 0 \text{ almost surely as } N \to \infty,
    \nonumber
\end{equation}
where $q_x^\gamma(x') := \frac{1}{\gamma}k(x-x')\cdot \frac{1}{K}$ with $K := \sum_{i=1}^N \frac{1}{\gamma} k(x-x_i)$ denotes the normalised distribution induced by a Borel kernel $k$ with the bandwidth satisfying $\gamma \to 0$ and $\frac{N \cdot \gamma^d}{\log N} \to \infty$ for $N \to \infty$.
\begin{proof}\ \\
We can reduce this problem to the consistency of kernel estimators; in the following, we adapt the notation used in \citep{Greblicki1984}, analogously to the proof of Proposition~\ref{prop:knn_consist}:
For fixed $\ph$, define $Z := \ph(X) - Y$.
Then $Z$ is bounded and we get a joint distribution of $(X,Z)$, a true calibration error $m(x) = \EE[Z|X=x] = \ph(x) - \ptr(x)$ and an estimator $m_N(x) := C(q_x^\gamma) = \sum_{i=1}^N q_x^\gamma(x_i) \cdot z_i$ with $z_i = \ph(x_i) - y_i$.\\
Additional constraints on $k$ are that
\begin{equation}
    c_1 \cdot H(\|x\|_s) \leq k(x) \leq c_2 \cdot H(\|x\|)
        \quad\text{and}\quad
    c_3 \cdot \mathbf{1}_{\|x\| \leq r}(x) \leq k(x)
    \nonumber
\end{equation}
for $c_1, c_2, c_3, r$ positive constants, and $H$ a bounded decreasing Borel function in $[0,\infty)$ such that
$t^d H(t) \to 0$ as $t \to \infty$.\\
Then by Corollary 1 from \citep{Greblicki1984}, we get 
$\EE \left[\left\vert m_N(x) - m(x) \right\vert\right] \to 0$ with probability 1 as $N \to \infty$, which is our desired result. 
\end{proof}

\ \\
For \textbf{Proposition~\ref{prop:quadrangle}} (restated below), we need some prerequisites.\\
The Quadrangle Theorem of \citep{Rockafellar2013} is stated in terms of two classes of agglomeration functions, for which we require the following further axiom for agglomeration functions $\Rb$ and $C \in \LL^2(J)$:
\ \\[3pt]
\begin{tabular}{lll}
\textbf{A7} &\textbf{Agreement}
 &\quad $C = \alpha \in \RR \text{ constant } \Rightarrow \Rb(C) = \alpha$.
\end{tabular}
\highlight{
\begin{definition}[Regular Risk Measures]\ \\
    A \emph{regular risk measure (RRM)} is an agglomeration function satisfying A7, Aversity and Convexity (cf. footnotes \ref{fn:convexity}, \ref{fn:closedness}).\label{def:RRM}
\end{definition}
}
\highlight{
\begin{definition}[Regular Deviation Measures]\ \\
    A \emph{regular deviation measure (RDM)} is an agglomeration function satisfying A6 and Convexity.\label{def:RDM}
\end{definition}
}

\noindent 
Now \citet{Rockafellar2013} establish that 
\begin{enumerate}[(A)]
    \item[(A)]
 A convex agglomeration function satisfies A2 if and only if it satisfies A7. 
    \item[(B)]
 Regular deviation measures $\Db$ stand in a one-to-one relationship with regular risk measures $\Rb$ via 
    \begin{equation}\label{eq:quadrangle2}
        \Rb(C) = \EE[C] + \Db(C).
    \end{equation}
    \item[(C)]
 $\Rb$ satisfies A3 if and only $\Db$ does. 
    \item[(D)]
 $\Rb$ satisfies A1 if and only if $\Db$ satisfies $\Db(C) \leq \sup C - \EE[C]\ \forall C \in \LL^2(J)$. 
\end{enumerate}
Furthermore, \citet{Artzner1999} establish that 
\begin{itemize}
    \item[(E)] An agglomeration satisfying A3 is convex if and only if it satisfies A4. 
\end{itemize}
Lastly, it is easy to see that
\begin{itemize}
    \item[(F)] For an RRM $\Rb$ and its corresponding RDM $\Db$, if one is law-invariant, then so is the other.
\end{itemize}
\ \\
\textbf{Proposition \ref{prop:quadrangle} (Relation between CRMs and FDMs).} \
    \begin{enumerate}[a)]
    \item[a)] Agglomeration functions $\Db$ satisfying A3-A6 stand in a one-to-one relationship with agglomeration functions $\Rb$ satisfying A2-A5 and Aversity (the property that $\Rb(C) > \EE[C]$ if $C \in \LL^2(J)$ is  not constant) via 
    \begin{equation}\label{eq:quadrangle3}
        \Rb(C) = \EE[C] + \Db(C).
    \end{equation}
    \item[b)] A law-invariant CRM induces an FDM via (\ref{eq:quadrangle3}) iff it is averse.
    \item[c)] An FDM induces a law-invariant CRM via (\ref{eq:quadrangle3}) iff it satisfies $\Db(C) \leq \sup C - \EE[C]\ \forall C \in \LL^2(J)$.
    \end{enumerate}
\begin{proof}\
\begin{enumerate}[a)]
\item[a)] An agglomeration function $\Db$ satisfying A3-A6 satisfies Convexity by (E) and is thus an RDM.
An agglomeration function $\Rb$ satisfying A2-A5 and Aversity satisfies A7 by (A) and Convexity by (E) and is thus an RRM.
The result follows from (B), (C), and (F).
\item[b)] This result follows directly from a) via the definitions of CRMs and FDMs.
\item[c)] This result follows from a) and (D).
\end{enumerate}
\end{proof}


\section{Partitions with Non-uniform Measures}
\label{app:partitions}

For groupings based on partitions, we may take group size into account in a different way compared to that in Sections \ref{s:scores} \& \ref{ss:fair_agglomeration} and Appendix~\ref{app:established_scores}:
Instead of counting groups multiple times in a grouping (as in (\ref{eq:aggr_ex_ece}) for the ECE), we may make use of the measure on the measure space $J$, call it $\mu$, that we have so far assumed to be uniform and thus left implicit (see the paragraph above Definition~\ref{def:agglom}).
This alternative formalisation, which encodes the information about group weights in the measure rather than in the multiplicities of the groups, has the advantage that we can treat partition-based groupings as partitions of the dataset since each datapoint is then a member of exactly one element of the grouping.
Using this alternative encoding of group weights, we provide an additional result that strengthens the choice of axioms in Sections \ref{s:scores} and \ref{ss:fair_agglomeration}, in particular Convexity (cf. footnote~\ref{fn:convexity}) and thereby also Subadditivity:
We show that the proposed global scores cannot decrease when making partitions \textit{finer}, that is, when further subdividing the groups (Definition~\ref{def:fine} below).
This can be seen as an extension of Proposition~\ref{prop:to_size_a} that also takes into account properties of the considered agglomeration functions.

\highlight{
\begin{definition}[Refinement, Finer, Coarser] \label{def:fine} \ \\
For two partitions $\PP_1, \PP_2$ of some set $\II$, the former is \emph{finer than} or \emph{a refinement of} the latter if every element of $\PP_1$ is a subset of some element of $\PP_2$.
This is equivalent to saying that $\PP_2$ is \emph{coarser} than $\PP_1$.
\end{definition}
}

\highlight{
\begin{proposition}[Finer partitions have higher scores] \label{prop:finer} \ \\
Take some $\ph: \XX \to [0,1]$, an agglomeration function $\Rb$, and a dataset $\{(x_i, y_i)\}_{i \in \II}$ with $\II := \{1,...,N\}$.
\begin{enumerate}[a)]
    \item[a)] If $\Rb$ is law-invariant and convex, plus if $\mu$ is given by the empirical distribution (that is, group errors are weighted by the size of the groups), then finer partitions cannot have lower global scores of raw (signed) errors than coarser ones. 
    
    \item[b)] If $\Rb$ is monotonic, law-invariant, and convex, plus if $\mu$ is given by the empirical distribution, then finer partitions cannot have lower global scores of absolute errors than coarser ones.
    
    
\end{enumerate}
\end{proposition}
}
\begin{proof}\
\begin{enumerate}[a)]
    \item[a)] 
    Take two partitions $\PP_1, \PP_2$ of $\II$ s.t. they contain the same sets except for some $I \in \PP_1$ and $I_k, I_l \in \PP_2$ with $I = I_k \cup I_l$.
    That means that $\PP_2$ is (slightly) finer than $\PP_1$ since $I \in \PP_1$ is further subdivided.
    Now let $J_1$ and $J_2$ denote the index sets of the partitions $\PP_1$ and $\PP_2$, equipped with the measure $\mu$ given by the empirical data distribution -- this means $\forall i \in \{1,2\}, j \in J_i: \mu(j) = \frac{|I_j|}{|\II|}$.
    We slightly abuse notation by overloading $\mu$ and $\Rb$ to denote the measures and agglomeration functions, respectively, on both $J_1$ and $J_2$.\footnote{Since $\mu$ is defined through the empirical distribution, the measure on $J_1$ and the one on $J_2$ are in some sense 'the same' despite operating on different spaces.
    Furthermore, to compare global scores on different partitions, we need some principle to identify agglomeration functions $\Rb$ operating on different measure spaces. 
    In the literature on inequality indices, this is done explicitly or implicitly through a shared functional form of inequality measures that can be applied to different population sizes.
    Here, this can be done even more naturally since agglomeration functions satisfying Law Invariance only depend on the \emph{distribution} of errors, not on the particularities of the measure spaces.
    Focusing on distributions as inputs, we can identify agglomeration functions on $J_1$ and on $J_2$ with each other and understand LI as a property of such functions operating on distributions.\label{fn:overload}}\\
    Let $C_1, C_2 \in \LL^2(J_1), \LL^2(J_2)$ denote the raw calibration errors of $\ph$ under $\PP_1, \PP_2$, respectively.
    Then 
    \begin{equation} \label{eq:finer_equal_avg}
        \frac{\mu(I_k)}{\mu(I)} c(I_k) + \frac{\mu(I_l)}{\mu(I)} c(I_l) = c(I)
    \end{equation}
    by definition of $c( \cdot )$ and $\mu$ (given $I = I_k \cup I_l$, similar to Proposition~\ref{prop:to_size_a}b).
    
    Now let $\tilde{C}_1 \in \LL^2(J_2)$ be equal to $C_2$ except that for $k,l \in J_2$ (corresponding to $I_k, I_l$), $\tilde{C}_1(k) = \tilde{C}_1(l) = c(I)$.
    That is, $\tilde{C}_1$ is defined on the same space as $C_2$ (namely $J_2$) while having the same distribution as $C_1$, given that $\frac{|I|}{|\II|} = \frac{|I_k| + |I_l|}{|\II|}$.
    Thus, by LI, $\Rb(C_1) = \Rb(\tilde{C}_1)$ (cf. footnote~\ref{fn:overload}).
    
    To prove $\Rb(C_1) \leq \Rb(C_2)$, it remains to be shown that $\Rb(\tilde{C}_1) \leq \Rb(C_2)$.

    By \citep{Bellini2021} Proposition~5.6, if $\Rb$ is LI and convex then it is \emph{dilatation monotone}, that is,
    \begin{equation}
        \Rb(\EE[C_2\ |\ \Gcal]) \leq \Rb(C_2)
    \end{equation}
    for any sub-$\sigma$-algebra $\Gcal$.
    This is precisely the property that we need, essentially meaning that by making a partition coarser, the agglomerated score may only decrease.
    
    Since in our case, ${C}_1$ ad $C_2$ are identical except that the former combines $I_k$ and $I_l$ into $I$ by averaging their scores as stated in (\ref{eq:finer_equal_avg}), we have $\tilde{C}_1 = \EE[C_2\ |\ \Gcal]$ for $\Gcal$ denoting the $\sigma$-algebra induced by the (coarser) partition $\PP_1$.
    Therefore, by Dilatation Monotonicity, $\Rb(\tilde{C}_1) \leq \Rb(C_2)$.

    Since we can represent any refinement of a partition on a finite set through steps where one set is split into two (as done here for $I = I_k \cup I_l$), the result holds for any two partitions where one is finer than the other.

    \item[b)] 
    Take the same $\PP_1, \PP_2, I, I_k, I_l, \mu$ as above.
    
    Now let $C_1, C_2 \in \LL^2(J_1), \LL^2(J_2)$ denote the \emph{absolute} calibration errors of $\ph$ under $\PP_1, \PP_2$, respectively.\\
    As in Proposition~\ref{prop:to_size_a}b), introducing $| \cdot |$ to (\ref{eq:finer_equal_avg}) yields
    \begin{equation} \label{eq:finer_higher_avg}
        \frac{\mu(I_k)}{\mu(I)} |c(I_k)| + \frac{\mu(I_l)}{\mu(I)} |c(I_l)| \geq |c(I)|.
    \end{equation}
    Define $\tilde{C}_1$ as above except that, given that we now consider absolute errors, $\tilde{C}_1(k) = \tilde{C}_1(l) = |c(I)|$.
    Then $\tilde{C}_1$ again has the same distribution as $C_1$ and thus, by LI, $\Rb(C_1) = \Rb(\tilde{C}_1)$.

    For the case of equality in (\ref{eq:finer_higher_avg}), we can show $\Rb(\tilde{C}_1) \leq \Rb(C_2)$ as in part a).
    
    For the case of strict inequality, define $C_3 \in \LL^2(J_2)$ to be equal to $C_2$ except that $C_3(k) < C_2(k)$ s.t. $\EE_\mu[C_3] = \EE_\mu[\tilde{C}_1]$.
    Then $\Rb(\tilde{C}_1) \leq \Rb(C_3)$ by Dilatation Monotonicity as in part a) and $\Rb(C_3) \leq \Rb(C_2)$ by Monotonicity.

    Thus, we again get $\Rb(C_1) \leq \Rb(C_2)$ in either case which proves the result.

    
    
    
\end{enumerate}
\end{proof}

\end{document}